\documentclass[twoside,11pt]{article}

\usepackage{jmlr2e}

\usepackage{microtype}
\usepackage{graphicx}
\usepackage{subfigure}
\usepackage{booktabs} 

\usepackage{hyperref}

\usepackage{multirow}
\usepackage{makecell}
\usepackage{color}
\usepackage{comment}
\usepackage{amsmath,amssymb} 
\usepackage{bbm}
\usepackage{adjustbox}
\usepackage{enumitem}
\usepackage{array}
\usepackage{caption}
\newcolumntype{$}{>{\global\let\currentrowstyle\relax}}
\newcolumntype{^}{>{\currentrowstyle}}
\newcommand{\rowstyle}[1]{\gdef\currentrowstyle{#1}%
  #1\ignorespaces
}
\usepackage[linesnumbered, ruled, vlined, algo2e]{algorithm2e}

\usepackage{amssymb}
\usepackage{amsmath,amsfonts}
\usepackage{amsopn,bm,mathtools}

\usepackage{nicefrac}       
\newcommand{\vct}[1]{\boldsymbol{#1}} 
\newcommand{\mat}[1]{\boldsymbol{#1}} 
\newcommand{\cst}[1]{\mathsf{#1}}  

\newcommand{\T}{^{\top}} 

\newcommand{\ProbOpr}[1]{\mathbb{#1}}

\newcommand{\expect}[2]{%
\ifthenelse{\equal{#2}{}}{\ProbOpr{E}_{#1}}
{\ifthenelse{\equal{#1}{}}{\ProbOpr{E}\left[#2\right]}{\ProbOpr{E}_{#1}\left[#2\right]}}} 
\newcommand{\var}[2]{%
\ifthenelse{\equal{#2}{}}{\ProbOpr{VAR}_{#1}}
{\ifthenelse{\equal{#1}{}}{\ProbOpr{VAR}\left[#2\right]}{\ProbOpr{VAR}_{#1}\left[#2\right]}}} 

\DeclareMathOperator{\argmin}{arg\,min}

\newcommand{\ones}{\vct{1}}

\newcommand{\va}{\vct{a}}

\newcommand{\ve}{\vct{e}}

\newcommand{\vg}{\vct{g}}

\newcommand{\vw}{\vct{w}}

\newcommand{\mS}{\mat{S}}

\newcommand{\mW}{\mat{W}}

\newcommand{\sD}{\mathcal{D}}

\newcommand{\sL}{\mathcal{L}}

\newcommand{\sN}{\mathcal{N}}

\newcommand{\vmu}{\vct{\mu}}
\newcommand{\mSigma}{\mat{\Sigma}}

\newcommand{\tp}{\tilde{p}}

\newcommand{\ttheta}{\tilde{\theta}}

\usepackage{xspace}

\makeatletter
\DeclareRobustCommand\onedot{\futurelet\@let@token\@onedot}
\def\@onedot{\ifx\@let@token.\else.\null\fi\xspace}
\def\eg{\emph{e.g}\onedot} 
\def\ie{\emph{i.e}\onedot}

\makeatother

\usepackage{mathtools}

\usepackage{listings}

\newcommand{\eat}[1]{}

\newcommand{\supp}{the Suppl. Material\xspace}

\newcommand{\ijknife}{infinitesimal jackknife\xspace}
\newcommand{\IJknife}{Infinitesimal jackknife\xspace}
\newcommand{\MLE}{\emph{\textsf{MLE}}\xspace}
\newcommand{\TCS}{\emph{\textsf{T. Scale}}\xspace}
\newcommand{\ENS}{\emph{\textsf{Deep E.}}\xspace}
\newcommand{\BNN}{\emph{\textsf{BNN}}\xspace}
\newcommand{\MCD}{\emph{\textsf{Dropout}}\xspace}
\newcommand{\BNNvi}{\emph{\textsf{BNN(VI)}}\xspace}
\newcommand{\KFAC}{\emph{\textsf{BNN(KFAC)}}\xspace}
\newcommand{\SWAG}{\emph{\textsf{SWAG}}\xspace}
\newcommand{\RUE}{\emph{\textsf{RUE}}\xspace}

\newcommand{\mfzero}{\emph{\textsf{mf0}}\xspace}
\newcommand{\mfone}{\emph{\textsf{mf1}}\xspace}
\newcommand{\mftwo}{\emph{\textsf{mf2}}\xspace}
\newcommand{\mfgs}{\emph{\textsf{mfGS}}\xspace}
\newcommand{\ukf}{\emph{\textsf{ukf}}\xspace}
\newcommand{\JK}{\emph{\textsf{JKnife}}\xspace}
\newcommand{\edl}{\emph{\textsf{EDL}}\xspace}
\newcommand{\Tens}{{T_{\cst{ens}}}}
\newcommand{\Tact}{{T_{\cst{act}}}}
\newcommand{\vomg}{\vct{\omega}}
\newcommand{\wh}{\widehat}
\newcommand{\htheta}{\wh{\theta}}

\usepackage{prettyref}
\newcommand{\pref}[1]{\prettyref{#1}}

\newcommand{\savehyperref}[2]{\texorpdfstring{\hyperref[#1]{#2}}{#2}}
\newrefformat{eq}{\savehyperref{#1}{\textup{(\ref*{#1})}}}
\newrefformat{eqn}{\savehyperref{#1}{Eqn.~\ref*{#1}}} 
\newrefformat{lem}{\savehyperref{#1}{Lemma~\ref*{#1}}}
\newrefformat{def}{\savehyperref{#1}{Definition~\ref*{#1}}}
\newrefformat{line}{\savehyperref{#1}{line~\ref*{#1}}}
\newrefformat{thm}{\savehyperref{#1}{Theorem~\ref*{#1}}}
\newrefformat{corr}{\savehyperref{#1}{Corollary~\ref*{#1}}}
\newrefformat{cor}{\savehyperref{#1}{Corollary~\ref*{#1}}}
\newrefformat{sec}{\savehyperref{#1}{\S~\ref*{#1}}}
\newrefformat{app}{\savehyperref{#1}{Appendix~\ref*{#1}}}
\newrefformat{assum}{\savehyperref{#1}{Assumption~\ref*{#1}}}
\newrefformat{ex}{\savehyperref{#1}{Example~\ref*{#1}}}
\newrefformat{fig}{\savehyperref{#1}{Figure~\ref*{#1}}}
\newrefformat{tab}{\savehyperref{#1}{Table~\ref*{#1}}}
\newrefformat{alg}{\savehyperref{#1}{Algorithm~\ref*{#1}}}
\newrefformat{rem}{\savehyperref{#1}{Remark~\ref*{#1}}}
\newrefformat{conj}{\savehyperref{#1}{Conjecture~\ref*{#1}}}
\newrefformat{prop}{\savehyperref{#1}{Proposition~\ref*{#1}}}
\newrefformat{proto}{\savehyperref{#1}{Protocol~\ref*{#1}}}
\newrefformat{prob}{\savehyperref{#1}{Problem~\ref*{#1}}}
\newrefformat{claim}{\savehyperref{#1}{Claim~\ref*{#1}}}
\newrefformat{que}{\savehyperref{#1}{Question~\ref*{#1}}}

\ShortHeadings{Gaussian-Softmax Integral for Uncertainty Estimation}{Lu, Ie, and Sha}
\firstpageno{1}

\begin{document}

\title{Mean-Field Approximation to Gaussian-Softmax Integral with Application to Uncertainty Estimation}

\author{\name Zhiyun Lu \email zhiyunlu@google.com \\
	\addr Google Inc. \\
	111 8th Avenue, \\
	New York, NY 10011 \\
	\AND
            \name Eugene Ie \email eugeneie@google.com \\
        \addr Google Research \\
        1600 Amphitheatre Pkwy, \\
       Mountain View, CA 94043 \\
       \AND
       \name Fei Sha \email fsha@google.com \\
       \addr Google Research \\
       1600 Amphitheatre Pkwy, \\
       Mountain View, CA 94043}

\editor{XXX}

\maketitle

\begin{abstract}
Many methods have been proposed to quantify the predictive uncertainty associated with the outputs of deep neural networks. Among them, ensemble methods often lead to state-of-the-art results, though they require modifications to the training procedures and are computationally costly for both training and inference. In this paper, we propose a new single-model based approach. The main idea is inspired by the observation that we can ``simulate'' an ensemble of models by drawing from a Gaussian distribution, with a form similar to those from the asymptotic normality theory, infinitesimal Jackknife, Laplacian approximation to Bayesian neural networks, and trajectories in stochastic gradient descents. However, instead of using each model in the ``ensemble'' to predict and then aggregating their predictions, we integrate the Gaussian distribution and the softmax outputs of the neural networks. We use a mean-field approximation formula to compute this analytically intractable integral. The proposed approach has several appealing properties: it functions as an ensemble without requiring multiple models, and it enables closed-form approximate inference  using only the first and second moments of the Gaussian. Empirically, the proposed approach performs competitively when compared to state-of-the-art methods, including deep ensembles, temperature scaling, dropout and Bayesian NNs, on standard uncertainty estimation tasks. It also outperforms many methods on out-of-distribution detection.

\end{abstract}


\section{Introduction}

Deep neural nets are known to output overconfident predictions.  For critical applications such as autonomous driving and medical diagnosis, it is thus necessary to quantify accurately the uncertainty around those predictive outputs.  As such, there has been a fast-growing interest in developing new methods for uncertainty estimation. 

In this paper, we consider feedforward neural nets for supervised learning of multi-way classifiers. We provide more discussions of (predictive) uncertainty estimation methods in \S\ref{sRelated} and cite a few  here.  They fall into three  categories: a single pass of a trained model to output both a categorical prediction and the associated probabilities~\citep{guo2017calibration},  multiple passes of an ensemble of different models so that the outputs are aggregated~\citep{lakshminarayanan2017simple}, and Bayesian approaches that output predictive posterior distributions~\citep{mackay1992bayesian}. The single-pass methods demand low on computation and memory. The ensemble approaches need to  train a few models, and also need to keep them around during test time, thus incurring higher computational costs and  memory requirement.   Bayesian neural networks are principled at incorporating prior knowledge though they are non-trivial to implement and the intractable posterior inference is often a serious challenge.  (Conceptually, we can also view Bayesian neural networks as an ensemble of an infinite number of models drawn from the posterior distribution.) While how to evaluate those techniques is an important research topic and has its own cautionary tales,  on many popular benchmark datasets, some of the ensemble approaches have consistently attained state-of-the-art performances~\citep{ashukha2020pitfalls}.

We study how we can achieve the competitive performance of ensemble methods while reducing their demand on computation and memory. Arguably, the competitive performance stems from aggregating (diverse) models.  Our main idea is thus inspired by the thought that we can pretend having an ensemble where the models are drawn from a multivariate Gaussian distribution. The mean and the covariance matrix of the distribution  describe the models succinctly. However, for neural networks used for classification, integrating this distribution with their \emph{softmax} output layers, so as to aggregate the models' outputs, is analytically intractable, as Bayesian posterior inference has  encountered.

We make two contributions. The first is to introduce a simple but effective approximation trick called mean-field Gaussian-Softmax~\citep{daunizeau2017semi} and extend it with several variations for uncertainty estimation. The method generalizes the Gaussian-sigmoid integral, a well-known approximation frequently used in machine learning~\citep{bishop06}. Likewise, it needs only the first and the second moments of Gaussian random variables and uses them in calculating the probabilistic  outputs. The resulting approximation scheme is also intuitive. In its simplest form, it provides an adaptive, data-dependent temperature scaling to the logits, analogous to the well-known approach of global temperature scaling~\citep{guo2017calibration}. 

Secondly, the specific form  (Eq.~(\ref{eGaussianVariance})) of the Gaussian distribution we assume for the \emph{pretended} ensemble is also very well connected to many ideas in the literature. Its form is the same as asymptotic normal distribution of parameters\footnote{We emphasis that this is just similar in superficial appearance.  The theory does not apply in its full generality since neural networks are not identifiable.}. It can be motivated from the infinitessimal jackknife technique~\citep{schulam2019can,giordano2018swiss,giordano2019higher,koh2017understanding}, where regularity conditions for asymptotic normality are relaxed. It can also be seen as the Laplace approximation of Bayesian neural networks~\citep{mackay1992bayesian,izmailov2018averaging,mandt2017stochastic}; or the sampling distribution of models collected on training trajectories~\citep{chen2016statistical,maddox2019simple}. Yet, despite the frequent appearance, to the best of our knowledge, none of those prior works has developed the proposed approximate inference approach for uncertainty quantification. In particular, the typical strategy resorts to \emph{sampling} from the Gaussian distribution and performing Monte Carlo approximation to the integral, a computation burden \emph{we explicitly want to reduce}.

We show combining these two lead to a method (\mfgs) that often surpasses or is as competitive as existing approaches in uncertainty quantification metrics and out-of-distribution detection on all benchmark datasets. It has several appealing properties:  constructing the Gaussian distribution does not add significant burden to existing training procedures; approximating the ensemble with a distribution removes the need of storing many models -- an impractical task for modern learning models. Due to its connection to Bayesian neural networks,  existing approaches in computational statistics such as Kronecker product factorization can be directly applied to accelerate computation and improve scalability.

We describe \mfgs in \S\ref{sMethod} and \S\ref{sUncertainty}, followed by a discussion on the related work in \S\ref{sRelated}. Empirical studies are reported in \S\ref{sExperiments}, and we conclude in \S\ref{sConclusion}. 
\section{Problem Setup}
\label{sSetup}
We are interested in developing a simple and effective way to compute approximately
\begin{equation}
e_k = \int \textsc{softmax}_k(\va(\theta)) \tp(\theta)d\theta,
\label{eMCEnsemble}
\end{equation}
where the $\textsc{softmax}_k(\cdot)$ is used to transform $\va(\theta)$ to a $K$-dimensional probabilistic output of a deep neural network parameterized with $\theta$. This integral is a recurring theme in Bayesian learning where the density $\tp(\theta)$ is the posterior distribution~\citep{bishop06}. The posterior distribution is interpreted as \emph{epistemic uncertainty} and the resulting $e_k$ is treated as the predictive uncertainty. This form also encompasses many ensemble methods where $\tp(\theta)$ is a discrete measure, representing each model (see below). The integral is not analytically tractable, even for ``simple'' distributions such as Gaussian.  As such, one often resorts to Monte Carlo sampling to approximate it. We seek to avoid that.

In the following, we introduce the notations and provide more detailed contexts that motivate this problem.

\paragraph{Notation} We focus on  (deep) feedforward neural network for $K$-way classification.  We are given a training set of $N$ \textit{i.i.d.}\! samples $\sD = \{z_i\}_{i=1}^n$, where $z_i = (x_i,y_i)$ with the input $x_i \in \mathcal{X}$  and the target $y_i \in \mathcal{Y} =\{1, 2,\cdots, K\}$. The output of the neural network is denoted by $o$. 

We optimize the model's parameter $\theta$ 
\begin{equation}
\htheta_n = \argmin_\theta \left(\sL(\sD; \theta) \stackrel{\textrm{def}}{=}\sum\nolimits_{i=1}^n \ell(z_i, o_i; \theta) \right),
\label{eLoss}
\end{equation}
where the loss is given by the sum of per-sample loss $\ell_i = \ell(z_i, o_i; \theta)$. In this work, we primarily focus on $\ell$ being the negative $\log$-likelihood, though other losses can also be considered~(\S\ref{sRationales}). The subscript $n$ signifies the number of training points -- a convention in mathematical statistics. See \S\ref{sRationales} for more examples.

We are interested in not only the categorical prediction on $x$'s class  but also the uncertainty of making such a prediction. We want the softmax probability $o_k$ to accurately represent the probability of belonging to the $k$-th class, 
\begin{equation}
o_k =  \textsc{softmax}_k(\va) = \frac{\exp\{a_k\}}{\sum_{k'}\exp\{a_{k'}\}} , 
\label{eSoftmax}
\end{equation}
where the activation (or logits) $a_k$ is  a linear transformation of the last-layer features $\vg(x; \phi_n)$, with the $k$-th column of the last layer parameter (matrix) $\mW_n$:
\begin{equation}
a_k = \vg(x; \phi_n)\T \mW_n(:, k).
\label{eLogits}
\end{equation}
In this notation, the neural network parameter $\theta$ is  the union of $\phi_n$ for all bottom layers and $\mW_n$ in the last layer.

\paragraph{Ensemble Methods} A classic yet very effective idea to improve the estimation Eq.~(\ref{eSoftmax}) is to use an ensemble of models. Let $\vartheta = \{\theta^{(1)}, \theta^{(2)}, \cdots, \theta^{(M)}\}$ denote the collection of $M$ models. The aggregated predictions is then
\begin{equation}
e_k = \frac{1}{M}\sum\nolimits_{m=1}^M \left(o_k^{(m)}\stackrel{\textrm{def}}{=}  \textsc{softmax}_k(\va^{(m)})\right),
\label{eEnsembleAverage}
\end{equation}
where the predictions (and activations) are computed using each model's parameters. However, obtaining and storing $M$ copies of parameters is a serious challenge in both computation and memory, especially for large models.  Besides, during inference it is necessary to do $M$ passes of different models to compute $e_k$.

Eq.~(\ref{eEnsembleAverage}) can be seen as the Monte Carlo approximation to Eq.~(\ref{eMCEnsemble}), if we treat $\vartheta$ as a sample drawn from a distribution of models $\tp(\theta)$. Now at least we do not have to keep the models around as the distribution implicitly ``memorizes'' the models.    In this perspective, the ensemble approach is clearly related to Bayesian neural networks (\BNN) when $\tp(\theta)$ is the posterior distribution $p(\theta | \sD)$ and the loss function $\sL(\sD; \theta)$ is based on a likelihood function and a prior on the model parameters.

However, there are still two issues to be resolved for the above conceptual framing to be practical. First, how do we choose $\tp(\theta)$?  Naturally we want to avoid as much as possible the cost of training multiple times and fit a distribution to the models. Ideally, \emph{we want to train one model and use it to derive the distribution.} 

Secondly, how do we make Eq.~(\ref{eMCEnsemble}) computationally tractable for the chosen $\tp(\theta)$?   Due to the nonlinearity of the softmax, unless $\tp(\theta)$ is a $\delta$ distribution or a discrete measure (\ie, the ensemble), the integral is not analytically tractable. Even if we have a way to sample from $\tp(\theta)$, resorting to sampling to compute the integral, as commonly used in Bayesian inference, will still make us face the same challenge of multiple passes of models as in Eq.~(\ref{eEnsembleAverage}).

We address these two points as follows: in \S\ref{sMethod}, we describe how to integrate Gaussian with softmax. In \S\ref{sUncertainty}, we describe how to use it for uncertainty estimation.

\section{Gaussian-Softmax Integral}
\label{sMethod}

Consider the following integral with respect to a $K$-dimensional Gaussian variable $\va \sim \sN(\va; \vmu, \mS)$,
\begin{equation}
e_k   = \int \textsc{softmax}_k(\va) \sN(\va; \vmu, \mS) \mathrm{d}\va. \label{eGaussianSoftmax}
\end{equation}

Eq.~(\ref{eGaussianSoftmax})  aggregates many classifiers's softmax outputs where $\va$ is each model's logits. This integral is not analytically tractable. We introduce a mean-field approximation scheme to compute it, without resorting to Monte Carlo sampling.

\subsection{Mean-Field Approximation}\label{sMF}

The main steps of the approximation scheme Eq.(\ref{eMF2})  in this subsection also appeared in~\citep{daunizeau2017semi}, though the author there did not use it or other forms Eq.(\ref{eMF0},\ref{eMF1}) for uncertainty estimation. Addiditionally, we discuss the intuitions of those approximations (\S\ref{sIntuition}), extend them with two temperatures for additional control (\S\ref{sUncertainty}), and showed that the simpler form of Eq.(\ref{eMF0}) is equally successful (\S\ref{sExperiments}).

The case $K=2$ is  well-known~\citep[p. 219]{bishop06} 
\begin{equation}
\int \sigma(a) \sN(a; \mu, s^2) \mathrm{d}a \approx \sigma\left(\frac{\mu}{\sqrt{1+\lambda_0 s^2}}\right),
\label{eGaussianSigmoid}
\end{equation}
where the softmax becomes the sigmoid function $\sigma(\cdot)$. $\lambda_0$ is a constant and is usually chosen to be $\pi/8$ or $3/\pi^2$.  To generalize to $K>2$, we rewrite the integral Eq.~(\ref{eGaussianSoftmax}) so that we compare $a_i$ to $a_k$ with $k \ne i$,  drawing close to Eq.~(\ref{eGaussianSigmoid}),
\begin{align}
e_k  &   = \int \frac{1}{1 + \sum_{i \ne k} \exp \{- (a_k - a_i)\} } \sN(\va; \vmu, \mS) \mathrm{d}\va
 \notag \\
& = \int \left( 2 -K + \sum_{i \ne k}\frac{1}{\sigma(a_k - a_i)} \right)^{-1} \sN(\va; \vmu, \mS) \mathrm{d}\va 
 \notag \\
&  \mathrel{\substack{\text{mean}\\\approx\\\text{field}}}   \left(2 -K + \sum_{i \ne k}\frac{1}{\expect{p(a_i, a_k)}{\sigma (a_k - a_i)}}\right)^{-1} ,  \label{eMFSepInt}
\end{align}
where ``mean field'' (MF)  ``pushes'' the integral through each sigmoid term independently\footnote{This approximation is prompted by the MF approximation $\expect{}{f(\cdot)} \approx f(\expect{}{\cdot})$ for nonlinear functions $f(\cdot)$. Moreover, we have factorized the $f()$ to be integrated as ``pairwise factors'', \ie $(a_k - a_i)$, and apply $p(a_i, a_k)$ to them -- this is the structural mean-field approach by factorizing joint distributions into ``smaller'' factors~\citep{koller2009}.}. Next we plug in the approximation Eq.~(\ref{eGaussianSigmoid})  to compute the inside expectations.  

\paragraph{Mean-Field 0 (\mfzero)}   We ignore the variance of $a_i$ for $i\ne k$ and replace $a_i$ with its mean $\mu_i$, and compute the expectation only with respect to $a_k$. We arrive at
\begin{align} 
e_k & \approx  \left(2 -K + \sum_{i \ne k}\frac{1}{\expect{p(a_k)}{\sigma (a_k - \mu_i)}}\right)^{-1} \notag \\
& \approx \left( 2-K + \sum_{i\ne k }\frac{1}{
\sigma\left( \frac{\mu_k - \mu_i}{\sqrt{1+ \lambda_0 s_k^2}}      \right)} \right)^{-1} \notag \\
& = \frac{1}{1 + \sum_{i \ne k} \exp\left\{- \frac{(\mu_k - \mu_i)}{\sqrt{1+ \lambda_0 s_k^2 }}\right\} } \notag \\
& = \textsc{softmax}_k\left(\frac{\vmu}{\sqrt{1+ \lambda_0 s_k^2 }}\right) \stackrel{\text{def}}{=}  \tilde{e}_k(0).
\label{eMF0}
\end{align}

\paragraph{Mean-Field 1 (\mfone)} If we replace $p(a_i, a_k)$ with the two independent marginals $p(a_i)p(a_k)$ in the denominator, recognizing $(a_k - a_i) \sim \sN(\mu_k-\mu_i, s_i^2+s_k^2)$, we get,
\begin{align}
e_k & \approx \left( 2-K + \sum_{i\ne k }\frac{1}{
\sigma\left( \frac{\mu_k - \mu_i}{\sqrt{1+ \lambda_0 (s_i^2 + s_k^2)}}     \right)} \right)^{-1}  \notag \\
& = \frac{1}{1 + \sum_{i \ne k} \exp\left\{- \frac{(\mu_k - \mu_i)}{\sqrt{1+ \lambda_0 (s_k^2+s_i^2) }}\right\} }  \stackrel{\text{def}}{=}  \tilde{e}_k(1).
\label{eMF1}
\end{align}

\paragraph{Mean-Field 2 (\mftwo)} Lastly, if we compute Eq.(\ref{eMFSepInt}) with a full covariance between $a_i$ and $a_k$, recognizing $(a_k - a_i) \sim \sN(\mu_k-\mu_i, s_i^2+s_k^2-2s_{ik})$, we get
\begin{align}
e_k  \approx \frac{1}{1 + \sum_{i \ne k} \exp\left\{- \frac{(\mu_k - \mu_i)}{\sqrt{1+ \lambda_0 (s_k^2+s_i^2-2s_{ik}) }}\right\} } \stackrel{\text{def}}{=}  \tilde{e}_k(2).
\label{eMF2}
\end{align}

Collectively, we refer to Eq.~(\ref{eMF0}-\ref{eMF2}) as mean-field Gaussian-Softmax (\mfgs).

\subsection{Intuition and Remarks}
\label{sIntuition}

Any of the 3 forms has approximation errors that depend on how ``nonlinear'' the integrand is. If one class is dominant over others, the softmax is similar to the sigmoid, where the approximation is well-known and applied in practice. We leave more precise quantification of those errors to future studies and focus in this work on their empirical utility. In particular,  the simple forms of \mfgs make it possible to understand them intuitively. 
 
First, note that when $\mS\rightarrow \mathbf{0}$, \ie, $p(\va)$ reduces to a $\delta$ distribution, \mfgs approximations become 
exact \begin{equation}
\tilde{e}_k(0), \tilde{e}_k(1), \tilde{e}_k(2) \rightarrow  \textsc{softmax}_k(\vmu).
\end{equation}
Namely, if there is no uncertainty associated with $\va$, the model should output the regular softmax --- this is a desirable sanity check of the approximation schemes. However, when $\va$ has variance, each  scheme handles it differently.

\mfzero is especially interesting.  The $\tilde{e}_k(0)$ is similar to the regular softmax with a temperature
\begin{equation}
T_k  = \sqrt{1+ \lambda_0 s_k^2}.
\end{equation}
However, as opposed to the regular temperature scaling factor~\citep{guo2017calibration}, $T_k$ is adaptive as it depends on the variance of a particular sample to be predicted.  If the activation on the $k$-th class has high variance, the temperature for that category is high, reducing the corresponding ``probability'' $\tilde{e}_k$. Specifically,
\begin{equation}
\tilde{e}_k \rightarrow \nicefrac{1}{K} \quad \text{as}\quad s_k \rightarrow +\infty.
\end{equation} 
In other words, the scaling factor is \emph{category and input-dependent}, providing additional flexibility to a global temperature scaling factor.

\mfone and \mftwo differ from \mfzero in how much information from $i\ne k$ is considered: the variance $s_i^2$ or additionally the covariance $s_{ik}$. The effect is to measure how much the $k$-th class should be confused with the $i$-th class after taking into consideration the variances in those classes.  Note that \mfzero and \mfone are computationally preferred over \mftwo which uses $K^2$ covariances, where $K$ is the number of classes.

\noindent
\textbf{Renormalization}\ 
Note that if Eq.~(\ref{eGaussianSoftmax}) is computed exactly, all $e_k$ should sum to 1. However, due to the category-specific scaling in the approximation schemes, $\tilde{e}_k$ is no longer a properly normalized probability.  Thus we renormalize as, 
\begin{equation}
e_k \approx \frac{\tilde{e}_k(0)}{\sum_i \tilde{e}_i(0)} \stackrel{\text{def}}{=}  \tp_k(0).
\label{eReNormalize}
\end{equation}
and similarly for \mfone and \mftwo too.


\section{Uncertainty Estimation}
\label{sUncertainty}
In this section, we describe how to apply \mfgs to the uncertainty estimation problem described in \S\ref{sSetup}. 

\subsection{The Form of $\tp(\theta)$}
We propose to use a Gaussian for $\tp(\theta)$ in Eq.~(\ref{eMCEnsemble}): 
\begin{equation}
\tp(\theta) = \sN(\theta;\, \htheta_n, {\Tens}\!^{-1}\mSigma_n),
\label{ePtheta}
\end{equation}
where $\htheta_n$ is the neural network parameter of the solution (Eq.~(\ref{eLoss})). $\Tens$ is a global ensemble temperature controlling the spread of the Gaussian by scaling the covariance matrix. The covariance $\mSigma_n$ is
\begin{equation}
\mSigma_n =  H_n^{-1}, J_n^{-1}, \text{or}\  H_n ^{-1}J_n H_n ^{-1},
\label{eGaussianVariance}
\end{equation}
where the Hessian $H_n$ and the variance of the gradients $J_n$ are defined as follows, computed at $\htheta_n$,
\begin{align}
H_n   =   \sum\nolimits_i \nabla\nabla\T\ell_i,\ J_n   = \sum\nolimits_i \nabla\ell_i \nabla \ell_i\T.
\label{eHessianFisher}
\end{align}
The Gaussian is used  not only because it is convenient but also the forms of $\mSigma_n$ have been occurring frequently in the  literature as ways to capture the variations in estimations of model parameters.  In \S\ref{sRationales}, we will discuss  the theoretical rationales behind the choices, after we describe how to use them for uncertainty estimation. (As a preview to our results, 
$\mSigma_n =  H_n^{-1}$ performs the best empirically.)

\subsection{Mean-field for Uncertainty Estimation}
\label{sMFUE}

To compute Eq.~(\ref{eMCEnsemble}), we make one final approximation, borrowing the linear perturbation idea from the influence function~\citep{cook1982residuals}, or infinitessimal jackknife, see Eq.~(\ref{eFuncPerturb}) in \S\ref{sRationales}. Consider the logits $\va(\theta)$ (Eq.~(\ref{eLogits})) when $\theta$ is close to $\htheta_n$, 
\begin{equation}
\va(\theta) \approx \va(\htheta_n) + \nabla \va\cdot (\theta-\htheta_n),
\label{eActLinear}
\end{equation}
where $\nabla \va$ is the gradient w.r.t. $\theta$.  Under $\theta \sim \tp(\theta)$, we have $\va(\theta)$ as a Gaussian as well,
\begin{align}
\va(\theta) \sim \sN(\va; \vmu= \va(\htheta_n), \mS   = \Tens^{-1}\nabla \va\mSigma_n\nabla \va\T)
\label{eActVariance}
\end{align}
Applying any form of \mfgs in \S\ref{sMethod} of computing Gaussian-softmax integral Eq.~(\ref{eGaussianSoftmax}), we  obtain a mean-field approximation to Eq.~(\ref{eMCEnsemble}). Note that the approximation in Eq.~(\ref{eActLinear})  is exact if $\va(\theta)$ is a linear function of the parameter, for instance, in the last-layer setting where all layers but the last are treated as  being deterministic.

\paragraph{Extension to ``Hard''-max} While it is common to think of softmax as outputting soft probabilities, it is also beneficial to introduce a global temperature scaling factor $\Tact$ so that we can adjust the ``softness'' for all the models in the ensemble. Specifically, we use $\textsc{softmax}_k(\Tact^{-1}\va)$. 

Note that   $\Tens$ and $\Tact$ control variability differently. 
When $\Tact \rightarrow 0$, each model in the ensemble moves to ``hard'' decisions. The ensemble average in Eq.~(\ref{eMCEnsemble}) thus approaches the majority voting.
When $\Tens \rightarrow \infty$, the ensemble focuses on one  model, cf. \citep{wenzel2020good}.  Empirically, we  tune the temperatures $\Tens$ and $\Tact$ as hyper-parameters on a heldout set, to optimize the predictive performance.

\textbf{Implementation} \pref{alg:uncert} illustrates the main steps of \mfgs to estimate uncertainty. Up to two passes through a trained network are needed. If we treat every layer but the last one as a deterministic mapping $g(x; \phi_n)$, we can eliminate the backward pass as the gradient of $\va(\theta)$ with respect to the last layer parameter $\mW_n$ is precisely  $g(x; \phi_n)$, already computed in the forward pass. As the empirical study in \S\ref{sExperiments} shows, this  setting already leads to very competitive or even the strongest performances.

For smaller models or the aforementioned last-layer setting, computing and storing $H^{-1}$  or $J^{-1}$ is advisable. For very large models, there are well-experimented tricks to approximate; we have used the Kronecker factorization satisfactorily and we only need to formulate those matrices for parameters in each layer independently~\citep{ritter2018scalable}. Others are possible~\citep{pearlmutter1994fast,agarwal2016second,koh2017understanding}. Note that the computation cost for those matrices are one-time and are amortized.

\begin{algorithm2e}[t]
\caption{Predictive Uncertainty Estimation with Mean-Field Gaussian-Softmax }\label{alg:uncert}
{\bf Input:} A deep neural net model $\htheta_n=(\phi_n, \mW_n)$, a training set $\sD$, $\Tens, \Tact$, and a test point $x$. \\
Compute $\mSigma_n$ (Eq.~(\ref{eGaussianVariance})) and cache it for future use\\
Forward Pass: compute the activation to the softmax layer $\vmu = \Tact^{-1} g(x; \phi_n)\T\mW_n$\quad \\
Backward Pass: compute the gradient $\nabla \va$\\
Compute the variance $\mS$ as in Eq.(\ref{eActVariance}))\\
Compute $\tilde{e}_k$ via mean-field approximation Eq.~(\ref{eMF0}), (\ref{eMF1}) or ~(\ref{eMF2}),  and then $\tp_k$, cf. Eq.~(\ref{eReNormalize}).
\\
\textbf{Output:} The predictive uncertainty is $\tp_k$.
\end{algorithm2e}

\subsection{Why Using the Gaussians in Eq.~(\ref{eGaussianVariance})?}\label{sRationales}
Readers who are familiar with  classical results in theoretical statistics and Bayesian statistics can easily identify the origins of the forms in Eq.~(\ref{eGaussianVariance}). An accessible introduction is provided in \citep{geyer2013}. Despite being widely known, they have not been applied extensively to deep learning uncertainty estimation. We summarize them briefly as follows. 

In the asymptotic normality theory of maximum likelihood estimation, for identifiable models, if the model is correctly specified and the true parameter is $\theta^*$,  the maximum likelihood estimation converges in distribution to a Gaussian
\begin{equation}
\sqrt{n}(\htheta_n - \theta^*) \stackrel{D}{\rightarrow}  \sN(0, I_1(\theta^*)^{-1}),
\end{equation}
where $I_1(\theta^*)$ is the Fisher information matrix, and equals to the variance of the gradients $J_1(\theta^*)$\footnote{The notation $V(\cdot)$ is used in \citep{geyer2013}. The subscript 1 refers to sample size 1. $I_n = n I_1$ and $J_n = n J_1$.}. In practice, the observed Fisher $J_1(\theta^*)$ is often used, and only requires the first-order derivatives:
\begin{equation}
\sqrt{n}(\htheta_n - \theta^*) \stackrel{D}{\rightarrow}  \sN(0, J_1(\theta^*)^{-1}).
\end{equation}
If the model is misspecified, the estimator converges to the so-called ``sandwich'' estimator
\begin{equation}
\sqrt{n}(\htheta_n - \theta^*) \stackrel{D}{\rightarrow}  \sN(0, I_1(\theta^*)^{-1}J_1(\theta^*)I_1(\theta^*)^{-1}).
\end{equation}
The true model parameter $\theta^*$ is unknown and so we use  a plug-in estimation such as $\htheta_n$, resulting  
 Eq.~(\ref{ePtheta}).
 
 Neural networks are not identifiable so the asymptotic normality theory does not apply. However, the following infinitessimal jackknife technique can also be used to motivate the use of those Gaussians.
 Jackknife is a well-known resampling method to estimate the confidence interval of an estimator~\citep{tukey1958bias,efron1981jackknife}. Each element $z_i$ is left out from the dataset $\sD$ to form a unique ``leave-one-out'' Jackknife sample $\sD_i = \sD - \{z_i\}$, giving rise to
\begin{equation}
\ttheta_i  = \argmin_\theta \sL(\sD_i; \theta).
\end{equation}
We obtain an ensemble of $n$ such samples $\{\ttheta_i\}_{i=1}^n$ and use them  to estimate the variances of $\htheta_n$ and the predictions made from $\htheta_n$. However, it is not feasible to retrain modern neural networks  $n$ times. \IJknife approximates the retraining~\citep{jaeckel1972infinitesimal,koh2017understanding,giordano2018swiss}. The basic idea is to approximate $\ttheta_i$ with a small perturbation to $\htheta_n$, \ie, \emph{infinitessimal jackknife}
\begin{equation}
\ttheta_i \approx \htheta_n + H_n^{-1} \nabla \ell_i ( \htheta_n ),
\end{equation}
where $H_n$ is the Hessian Eq.~(\ref{eHessianFisher}). We fit the $n$ copies of $\ttheta_i$ with  a Gaussian distribution
\begin{equation}
\ttheta_i \sim \sN\left(\htheta_n, \frac{1}{n} H_n^{-1}J_nH_n^{-1}\right),
\end{equation}
which is very similar to the asymptotic normality results. 
The perturbation idea can also be used to approximate any differentiable function $f(\ttheta_i)$, \eg the logits in Eq.~(\ref{eActLinear}).
\begin{equation}
f(\ttheta_i) \approx f(\htheta_n ) + \nabla f\T (\ttheta_i - \htheta_n ).
\label{eFuncPerturb} 
\end{equation}
If we  extend the above procedure and analysis to bootstrapping (\ie, sampling with replacement), we will arrive at the following Gaussian to characterize infinitessimal bootstraps:
\begin{equation}
\ttheta_i \sim \sN\left(\htheta_n, H_n^{-1}J_nH_n^{-1} \right).
\end{equation}
More details can be found in ~\citep{giordano2018swiss}. Finally, the Laplace approximation in Bayesian neural networks uses
$\theta \sim \sN(\theta^*, H^{-1})$
to approximate the posterior distribution  where $\theta^*$ is the maximum-a-posterior estimate. 

In short, all those Gaussian approximations to the true distribution of $\htheta_n$ (\ie the epistemic uncertainty) share the similar form as Eq.~(\ref{eGaussianVariance}) with an added temperature $\Tens$ controlling the scaling factor (such as $\nicefrac{1}{n}$).

\section{Related Work}
\label{sRelated}

Ensemble methods have a long history in statistics and machine learning. Despite their effectiveness and often superior performance in uncertainty quantification, it is challenging to apply them to deep learning  as they require training multiple copies of models and keeping the models in the memory during testing time. 

Obtaining a diverse set of models to aggregate is important. This involves changing training strategy: random  initialization and hyperparameters~\citep{lakshminarayanan2017simple,ashukha2020pitfalls}, using different data partitions such as resampling or bootstraps~\citep{schulam2019can,efron1992bootstrap}, collecting models on training trajectories~\citep{chen2014stochastic, chen2016statistical,maddox2019simple,huang2017snapshot}. For Bayesian neural networks (\BNN) as an ensemble of an infinite number of models, it is important also for sampling or variational inference to aggregate predictions from models drawn from the posterior distributions~\citep{barber1998ensemble,gal2016dropout,blundell2015weight,ritter2018scalable,mackay1992bayesian,wenzel2020good}. 

Our work is inspired by two observations. Classical results from theoretical statistics describe the convergence of a series of models asymptotically to normal distributions~\citep{geyer2013,giordano2018swiss,giordano2019higher}. Yet, using such models as an ensemble, or equivalently, the distribution, for uncertainty estimation is under-explored. Secondly, real-world applications would require us to compute the Gaussian-Softmax integral approximately without resorting to sampling to lower the computational cost.  While  approximating for Gaussian-sigmoid integral is a well-known result for Bayesian logistic regression,  extending it to softmax is less common. Our empirical studies suggest its outstanding utility in matching or exceeding state-of-the-art methods.

Last but not least, recent work in evidential deep learning or prior networks treat the outputs of neural nets as samples of a Dirichlet distribution~\citep{sensoy18evidential,malinin18prior,hobbhahn2020fast}. They are also single-pass methods  and have shown promising results and we compare to one of them in the empirical studies.

\section{Experiments}
\label{sExperiments}

We first describe the setup. We then analyze the effectiveness of the \mfgs approximation schemes. We then provide a detailed comparison to  state-of-the-art and other popular approaches for  uncertainty estimation.  
\subsection{Setup}  

We provide key information in the following. For more details, please see \supp.

\noindent
\textbf{Evaluation Tasks and Metrics}\ We evaluate on two tasks: predictive uncertainty on in-domain samples, and detection of out-of-distribution samples. For in-domain predictive uncertainty, we report classification error rate $\varepsilon$ ($\%$), negative log-likelihood (NLL), and expected calibration error in $\ell_1$ distance (ECE $\%$) on the test set~\citep{guo2017calibration}. 

NLL measures the KL-divergence between the empirical label distribution and the predictive distribution of the classifiers. ECE measures the discrepancy between the histogram of the predicted probabilities by the classifiers and the observed ones in the data -- properly calibrated classifiers should yield matching histograms. Both metrics are commonly used in the literature and the lower the better.

On the task of out-of-distribution  (OOD) detection, we assess how well $\tp(x)$, the classifier's output being interpreted as probability, can be used to distinguish invalid samples from normal in-domain ones. Following the common practice~\citep{hendrycks2016baseline,liang2017enhancing,lee2018simple}, we report two threshold-independent metrics: area under the receiver operating characteristic curve (AUROC), and area under the precision-recall curve (AUPR). Since the precision-recall curve is sensitive to the choice of positive class, we report both ``AUPR in:out'' where in-distribution and out-of-distribution samples are swapped. We also report detection accuracy, the optimal accuracy achieved among all thresholds in classifying in-/out-domain samples. All three metrics are the higher the better.  

\noindent
\textbf{Datasets}\ For in-domain tasks, we use 4 image datasets: MNIST, CIFAR-10, CIFAR-100, and ImageNet (ILSVRC-2012)~\citep{lecun1998mnist,krizhevsky2009learning,deng2009imagenet}. For out-of-domain detection, another 4 datasets are used: NotMNIST, LSUN (resized), SVHN and ImageNet-O~\citep{bulatov2011notmnist,yu2015lsun,netzer2011reading,hendrycks2019natural}. In \supp, we list their descriptive statistics. In short, all but ImageNet have about 50K images for training, 1K to 5K for validation and about 10K images for testing, in 10 to 100 classes. ImageNet is significantly larger:  about 1.2 million, 25K and 25K for training, validation and testing, in 1000 classes.

\textbf{Classifiers}\ For MNIST, we train  a two-layer MLP with 256 ReLU units  per layer, using Adam optimizer for 100 epochs. For CIFAR-10, we train a ResNet-20 with Adam for 200 epochs. On CIFAR-100, we train a DenseNet-BC-121 with SGD optimizer for 300 epochs. For ImageNet, we train a ResNet-50 with SGD optimizer for 90 epochs.

\noindent
\textbf{Methods to Compare}\  We compare to state-of-the-art and popular approaches: the point estimate of maximum likelihood estimator (\MLE), the temperature scaling calibration (\TCS)~\citep{guo2017calibration}, evidential deep learning (\edl) or prior networks~\citep{sensoy18evidential,malinin18prior}, the deep ensemble method (\ENS)~\citep{lakshminarayanan2017simple}, Jackknife (\JK)~\citep{efron1981jackknife}, and the resampling infinitessimal bootstrap \RUE~\citep{schulam2019can}. For variants of Bayesian neural networks (\BNN), we compare to  \MCD~\citep{gal2016dropout},  \BNNvi, \KFAC~\citep{ritter2018scalable}, and \SWAG~\citep{maddox2019simple}.

\noindent
\textbf{Tuning}\ We use the NLL and AUROC on the held-out sets to tune hyper-parameters $\Tact$ and $\Tens$ for in-domain and out-domain tasks respectively. All methods use the same classifier architecture.  We also follow the published guidelines on tuning methods we compare to. For \edl, we use the publicly available code~\citep{edlpubliccode} and tuned accordingly. We will make public the implementation of our method (best temperatures are in \supp).

\begin{table}[t]
\caption{\mfgs outperforms Monte Carlo sampling for  Gaussian-Softmax integral.  $H^{-1}$ performs the best as the Gaussian covariance $\mSigma_n$.}
\label{tMFGS}
\centering
\small
\setlength{\tabcolsep}{3pt}
\begin{tabular}{cccc||ccc}
\hline
 &  \multicolumn{3}{c||}{MNIST } & \multicolumn{3}{c}{NotMNIST } \\ 
 & \multicolumn{3}{c||}{In-domain ($\downarrow)$ } & \multicolumn{3}{c}{OOD detection ($\uparrow$)}\\
\cline{2-7}
& $\varepsilon$ & NLL & ECE  & Acc. & ROC & PR(in : out) \\ \hline
\multicolumn{7}{l}{\mfgs \emph{with $H^{-1}$ versus Sampling with $M$ samples }}\\ \hline
$M=20$ & 1.7 & 0.06 & 0.4  &   87.6 &  93.5 &  91.7 : 94.3 \\
$M=100$ & 1.7 & 0.05 & 0.5 &  90.1 &  95.6 &  94.6 : 96.1 \\
$M=500$ & 1.7 & 0.05 & 0.5 & 90.8 &  96.2 &  95.8 : 96.5  \\  \hline
\mfzero & 1.7 & \bf{0.05} & \bf{0.2} &  \bf{91.9} &  \bf{96.9} & \bf{96.7 : 97.0}  \\ 
\mfone & 1.7 & 0.05 & 0.5 &  91.9 &  96.9 &  96.7 : 97.0  \\
\mftwo & 1.7 & 0.05 & 0.5 & 91.9 &  96.9 &   96.7 : 97.0  \\ \hline \hline
\multicolumn{7}{l}{\emph{mean-field} \mfzero \emph{with different covariances}} \\ \hline
$J^{-1}$ & 1.7 & 0.05 & 0.2 & 90.8 & 96.1 & 95.9 : 96.2 \\ 
$H^{-1}JH^{-1}$ & 1.7 & 0.05 & 0.3 & 87.0 & 93.4 & 92.1 : 94.0 \\ 
\SWAG &  1.6 & 0.05 & 0.3 & 87.6 & 93.3 & 91.5 : 94.0 \\
\hline
\end{tabular}
\vspace{-1em}
\end{table}

\begin{table}[t]
\caption{\mfgs with Different Number of Layers.}
\label{tLayers}
\centering
\small
\setlength{\tabcolsep}{3pt}
\begin{tabular}{rccc||ccc}
\hline
 \multirow{3}{*}{ Layers}  &  \multicolumn{3}{c||}{MNIST } & \multicolumn{3}{c}{NotMNIST } \\ 
 & \multicolumn{3}{c||}{In-domain ($\downarrow)$ } & \multicolumn{3}{c}{OOD detection ($\uparrow$)}\\
\cline{2-7}
& $\varepsilon$ & NLL & ECE  & Acc. & ROC & PR(in:out) \\ \hline
\multicolumn{7}{l}{Sampling with $M=500$ samples \emph{with different layers. }}\\ \hline
last layer & 1.7 & 0.05 & 0.5 				& 90.8 &  96.2 &  95.8 : 96.5  \\  
top two layers & 1.7 & 0.05 & 0.5 			&   90.9 & 96.2 & 96.0 : 96.4 \\ 
all three layers &  1.7 & 0.06 & 0.4 			& 87.5 & 92.5 & 87.0 : 93.5\\ \hline \hline
\multicolumn{7}{l}{\mfzero \emph{with different layers.}} \\ \hline
last layer &  1.7 & 0.05 & 0.2 				&  91.9 &  96.9 & 96.7 : 97.0  \\ 
top two layers &  1.7 & 0.05 & 0.3 			& 91.9 & 96.9 & 96.7 : 97.0 \\ 
all three layers & 1.7 & 0.05 & 0.3 			&  91.9 & 96.9 & 96.8 : 97.0\\
\hline
\end{tabular}
\vspace{-1em}
\end{table}

\begin{table*}[t]
\centering
\caption{\mfgs for Predictive Uncertainty Estimation on Large-scale Benchmark Datasets.} \label{tOtherDatasets}
\small
\setlength{\tabcolsep}{3pt}
\adjustbox{max width=\textwidth}{
\begin{tabular}{c  ccc |ccc  ||ccc|ccc}
\hline
\multirow{2}{*}{ Method}  & \multicolumn{3}{c|}{CIFAR-100} & \multicolumn{3}{c||}{ImageNet} & \multicolumn{3}{c|}{CIFAR-100 versus LSUN} & \multicolumn{3}{c}{ImageNet versus ImageNet-O}\\ \cline{2-13}
  &      $\varepsilon$ (\%)& NLL & ECE (\%) & $\varepsilon$ (\%)& NLL & ECE (\%) &  Acc. & ROC & PR (in:out)   &  Acc. & ROC & PR (in:out)   \\ \hline
\ENS (5)	 &  \bf{19.6} & \bf{0.71} & 2.00 	    & \bf{21.2} &   \bf{0.83 }&   3.10  & 74.4 &  82.3 &  85.7 : 77.8 & 60.1 &  50.6 &  78.9 : 25.7\\ 
\TCS  	 &  24.3 & 0.92 & 3.13 		& 23.7 &   0.92 &   2.09 & 76.6 &  84.3 &  86.7 : 80.3 & 58.5 &  54.5 &  79.2 : 27.8 \\ 
\KFAC	 & 24.1 & 0.89 & 3.36 		 &23.6 &   0.92 &   2.95 & 72.9 &  80.4 &  83.9 : 75.5 &  60.3 &  53.1 &  79.6 : 26.9  \\  \hline
\mfzero	 &  24.3 & 0.91 & \bf{1.49}  & 23.7 &   0.91 & \bf{0.93}  & \bf{82.2} &  \bf{89.9} & {\bf{92.0}} : {\bf{86.6}} & \bf{63.2} &   \bf{62.9} &   {\bf{83.5}} : {\bf{33.3}}\\ \hline
\mfone &  24.3 & 0.91 & 2.20 & 23.7 & 0.91 & 1.17 & 82.2 & 89.9 & 92.0 : 86.6 & 63.2 & 62.9 & 83.5 : 33.3 \\ \hline
\mftwo & 24.3 & 0.91 & 2.22 & 23.7 & 0.91 & 1.17 &  82.2 & 89.9 & 92.0 : 86.6 & 63.2 & 62.9 & 83.5 : 33.3 \\ \hline

\end{tabular}
}
\end{table*}

\begin{table}[t]
\caption{\mfgs Compared with SOTA Methods.}\label{tMFOthers}
\centering
\small
\setlength{\tabcolsep}{3pt}
\begin{tabular}{cccc||ccc}
\hline
 &  \multicolumn{3}{c||}{MNIST } & \multicolumn{3}{c}{NotMNIST } \\ 
 & \multicolumn{3}{c||}{In-domain ($\downarrow)$ } & \multicolumn{3}{c}{OOD detection ($\uparrow$)}\\
\cline{2-7}
 & $\varepsilon$ & NLL & ECE  & Acc. & ROC & PR(in : out) \\ \hline
\ENS(5)  & 1.3 & 0.05 & 0.3    & 86.5 &  88.0 &  70.4 : 92.8  \\ 
\ENS(500) & \bf{1.2} & \bf{0.04} & \bf{0.2} & 92.9 & 96.1 & 90.6 : 96.6\\ 
\JK(500) & \bf{1.2} & \bf{0.04} & 0.3 & \bf{93.4} & 96.6 & 91.8 : {\bf{97.0}}\\ 
\RUE 		& 1.7 & 0.08 & 0.9 & 61.1 &  64.7 &  60.5 : 68.4 \\ \hline
\MLE 			& 1.7 & 0.10 & 1.2 &   67.6 &  53.8 & 40.1 : 72.5\\
\TCS  	    	& 1.7 & 0.06 & 0.7 & 67.4 &  66.7 &  48.8 : 77.0 \\ 
\edl		&  1.8 & 0.09 & 1.1 &  83.1 & 80.9 & 59.8 : 90.1\\ \hline
\MCD			& 1.7 & 0.06 & 0.7 &  88.8 &  91.4 &  78.7 : 93.5 \\ 
\BNNvi     & 1.7 & 0.14 & 1.1  &  86.9 &  81.1 &  59.8 : 89.9\\ 
\KFAC	        & 1.7 & 0.06 &  \bf{0.2}  &  88.7 &  93.5 &  89.1 : 93.8\\ \hline
\mfzero 	& 1.7 & 0.05 & \bf{0.2} &  91.9 &  \bf{96.9} & {\bf{ 96.7}} : {\bf{97.0}}  \\ \hline
\end{tabular}
\vspace{-1em}
\end{table}

\subsection{How Good is Mean-field Approximation?}

We validate \mfgs  by comparing it to Monte Carlo sampling for computing the integral Eq.~(\ref{eGaussianSoftmax}). Since there is no ground-truth to this integral, we validate its utility in estimating predictive uncertainty as a ``downstream task''. To this end, we use the MNIST and NotMNIST datasets and perform uncertainty estimation with only the last layer. 

In Table~\ref{tMFGS}, results from all mean-field approximation schemes are better than those from sampling, in both in-domain and out-domain metrics. Note that while sampling could be improved by increasing samples (thus,  more computational costs and memory demands),  the mean-field approaches still manage to outperform, especially on the OOD metrics, probably because sampling requires a significantly more samples to converge (the observed improvement is at a fairly slow rate, preventing us from trying more samples.).

Among the 3 schemes, \mfzero performs better in ECE, an in-domain task metric while being identical to others.  This might be an interesting observation worthy of further investigation. One possible explanation is that for \emph{OOD samples}, the variances of the logits are such that $s_i \approx s_k \gg s_{ik}$, \ie $\mS = g(x;\phi_n) H_n^{-1} g(x;\phi_n)\T$\footnote{Eq.~(\ref{eActVariance}) after proper vectorization and padding.} is diagonally dominant -- the features of an OOD $x$ is in the null space of $H$. This is intuitive as $H$ should contain vectors maximizing inner products with features from the \emph{in-domain} samples.

Among 3 choices of selecting $\mSigma_n$ in Eq.~(\ref{eGaussianVariance}),  $H^{-1}$ has a clear advantage over the other two $J^{-1}$ and $H^{-1}JH^{-1}$, though $J^{-1}$ is reasonably good. We also experimented with the covariance matrix derived from the samples on the training trajectories, \ie \SWAG. \SWAG performs similar as \mfzero with $H^{-1}JH^{-1}$. This is  re-assuring as it confirms the theoretical analysis that \SWAG converges to the sandwich estimator when its trajectory is in the vicinity of the MLE~\citep{maddox2019simple}.

In short, the study suggests the \mfgs approximation has high quality, measured in the metrics of the ``downstream'' task of uncertainty estimation. Thus, in the rest of the paper, we focus on \mfzero using $H^{-1}$ as $\mSigma_n$.

  \subsection{Use More Layers for Uncertainty Estimation} 
In this section, we study the effect of using different number of layers of parameters in \mfgs for uncertainty estimation. We report both sampling and \mfzero results in Table~\ref{tLayers}.

\mfzero consistently outperforms Monte Carlo sampling. We also observe that using more layers than the last layer does not further help. This is in line with recent studies~\citep{zeng2018relevance,kristiadi2020being} showing that last-layer Bayesian approximation can yield good performances in uncertainty tasks. Thus we focus on \mfgs with the last layer for the rest of uncertainty estimation experiments.

\subsection{\mfgs for Uncertainty Estimation }

\noindent
\textbf{Detailed Comparison on (Not-)MNIST}\ Table~\ref{tMFOthers} contrasts \mfzero to ensemble methods, single-pass methods and \BNN methods. The ensembles of 5 or 500 models, either with the randomly initialized training (\ENS) or with \JK\footnote{We do leave-$k$-out training where $k=500$.} clearly dominate in terms of in-domain metrics, closely followed by \mfzero and \KFAC. However, in  OOD metrics, \mfzero clearly dominates others or is on par with the ensemble of 500 \JK models. 
\RUE does not perform well. We believe its infinitesimal bootstraps, which samples from training examples does not capture the full variations of model parameters, though it does perform better than \MLE. Note that other single-pass methods perform the worst. 

\citeauthor{ashukha2020pitfalls} suggest to assess new uncertainty estimate methods by checking  the number of models in \ENS needed to have a matching performance~\citep{ashukha2020pitfalls}. 
In OOD metrics, \mfzero is about 500 models in \ENS, representing a significant reduction in computational cost and memory requirement --- it took about 30 hours on a TitanX GPU to train \ENS (500) while \mfzero needs only one model and a few seconds of inference time.  In in-domain metrics, \mfzero needs to be improved in accuracy, which is left as future work. 

\noindent
\textbf{Results on Large-scale Datasets}\ Table~\ref{tOtherDatasets} extends the  comparison on other benchmark datasets, focusing on a few representative and competitive approaches (as seen from Table~\ref{tMFOthers}). \mfzero  dominates all other approaches in OOD metrics by a visible margin. On in-domain task, \mfzero falls behind \ENS on error rates and slightly on NLLs, but significantly improves on ECE. Due to the compute cost, \ENS used 5 models, as suggested in the original paper~\citep{lakshminarayanan2017simple}. Additionally, \mfzero performs strongly better than either the single-model or the \BNN methods in ECE.

\noindent
\textbf{More  Results in \supp}
Reported there are a fuller Table~\ref{tOtherDatasets}  from other approaches and the dataset CIFAR-10, as well as the analysis of the temperatures $\Tens$ and $\Tact$.

\section{Conclusion}\label{sConclusion}
We propose using mean-field approximation  to Gaussian-Softmax integral for uncertainty estimation. This is enabled by ``ensembling'' the models from a Gaussian distribution whose mean and covariance are readily derived from a trained single model. The proposed approach is compared to many state-of-the-art methods. We find it perform the strongest on identifying out-of-domain samples and is on par with other approaches on in-domain tasks. The method is appealing for its simplicity and effectiveness.


\acks{Fei Sha is on leave from University of Southern California. This work is partially supported by NSF
Awards IIS-1513966/1632803/1833137, CCF-1139148, DARPA Award\#: FA8750-18-2-0117, FA8750-19-1-
0504, DARPA-D3M - Award UCB-00009528, Google Research Awards, gifts from Facebook and Netflix, and
ARO\# W911NF-12-1-0241 and W911NF-15-1-0484.}


\newpage

\appendix

\section{More Experiment Results}
\subsection{An Extended Table 3 of the Main Text}
Table~\ref{tExtendTable2} provides an extended version of Table 3 in the main text to include results from more approaches. \mfzero outperforms all other approaches in OOD metrics including ensemble methods, single-pass methods, and \BNN methods. For in-domain tasks, \mfzero performs better than either the single-model or  \BNN methods in ECE. Ensemble performs the strongest in terms of accuracy and NLL.
\begin{table*}[th]
\centering
\caption{\mfgs for Predictive Uncertainty Estimation on Large-scale Benchmark Datasets.} \label{tExtendTable2}
\small
\setlength{\tabcolsep}{3pt}
\adjustbox{max width=\textwidth}{
\begin{tabular}{c  ccc |ccc  ||ccc|ccc}
\hline
\multirow{2}{*}{ Method}  & \multicolumn{3}{c|}{CIFAR-100} & \multicolumn{3}{c||}{ImageNet} & \multicolumn{3}{c|}{CIFAR-100 versus LSUN} & \multicolumn{3}{c}{ImageNet versus ImageNet-O}\\ \cline{2-13}
  &      $\varepsilon$ (\%)& NLL & ECE (\%) & $\varepsilon$ (\%)& NLL & ECE (\%) &  Acc. & ROC & PR (in:out)   &  Acc. & ROC & PR (in:out)   \\ \hline
\ENS (5)	 &  \bf{19.6} & \bf{0.71} & 2.00 	    & \bf{21.2} &   \bf{0.83 }&   3.10  & 74.4 &  82.3 &  85.7 : 77.8 & 60.1 &  50.6 &  78.9 : 25.7\\ 
\RUE$^\dagger$     &  24.3 & 0.99 & 8.60 		& 23.6 & 0.92 &  2.83 & 75.2 & 83.0 &  86.7 : 77.8 & 58.4 &  51.6 &  78.3 : 26.3  \\  \hline 
\MLE 	 &  24.3 & 1.03 & 10.4 		& 23.7&   0.92 &   3.03 & 72.7 &  80.0 &  83.5 : 75.2 &  58.4 &  51.6 &  78.4 : 26.3\\
\TCS  	 &  24.3 & 0.92 & 3.13 		& 23.7 &   0.92 &   2.09 & 76.6 &  84.3 &  86.7 : 80.3 & 58.5 &  54.5 &  79.2 : 27.8 \\   \hline
\MCD	 &  23.7 	& 0.84 & 3.43 	 & 24.9 &   0.99 &   1.62 & 69.8 &  77.3 &  81.1 : 72.9 &  59.5 &  51.7 &  79.0 : 26.3 \\ 
\BNNvi$^\dagger$   &  25.6 	& 0.98 & 8.35 	 & 26.5 &   1.17 &   4.41 & 62.5 &  67.7 &  71.4 : 63.1 &  57.8 &  52.0 &  75.7 : 26.8   \\ 
\KFAC	 & 24.1 & 0.89 & 3.36 		 &23.6 &   0.92 &   2.95 & 72.9 &  80.4 &  83.9 : 75.5 &  60.3 &  53.1 &  79.6 : 26.9  \\  \hline
\mfzero	 &  24.3 & 0.91 & \bf{1.49}  & 23.7 &   0.91 & \bf{0.93}  & \bf{82.2} &  \bf{89.9} & {\bf{92.0}} : {\bf{86.6}} & \bf{63.2} &   \bf{62.9} &   {\bf{83.5}} : {\bf{33.3}}\\ 
\mfone &  24.3 & 0.91 & 2.20 & 23.7 & 0.91 & 1.17 & 82.2 & 89.9 & 92.0 : 86.6 & 63.2 & 62.9 & 83.5 : 33.3 \\ 
\mftwo & 24.3 & 0.91 & 2.22 & 23.7 & 0.91 & 1.17 &  82.2 & 89.9 & 92.0 : 86.6 & 63.2 & 62.9 & 83.5 : 33.3 \\ \hline
\end{tabular}
}

{$^\dagger$: Only the last layer is used  due to the high computational cost.}
\end{table*}

\subsection{More Results on CIFAR-10 and CIFAR-100}~\label{sCIFARmore}
Tables~\ref{tab:tCifar10} and \ref{tab:tCifar100svhn} supplement the main text with additional experimental results on the CIFAR-10 dataset with both in-domain and out-of-distribution detection tasks, and on the CIFAR-100 with out-of-distribution detection using the SVHN dataset. \emph{\textsf{BNN(LL-VI)}}  refers to \BNN with stochastic variational inference applied on the last layer only, following~\cite{snoek2019can}.

The results support the same observations in the main text:  \mfzero noticeably outperforms other approaches on out-of-distribution detection, but is not as strong as ensemble method in terms of accuracy and NLL. It does in general outperform all other approaches in those two metrics.

\begin{table*}[t]
\centering
\caption{\mfgs for Predictive Uncertainty Estimation on CIFAR-10.}\label{tab:tCifar10}
\begin{adjustbox}{max width=\textwidth}
\begin{tabular}{$c^c^c^c||^c^c^c|^c^c^c}
\hline
\multirow{2}{*}{Method} & \multicolumn{3}{c||}{CIFAR-10} 
& \multicolumn{3}{c|}{CIFAR-10 verus LSUN} 
& \multicolumn{3}{c}{CIFAR-10 verus SVHN}\\ \cline{2-10}
& $\varepsilon$ (\%)& NLL & ECE (\%) & Acc. & ROC & PR (in:out) & Acc. & ROC & PR (in:out)\\ \hline	 
\ENS	 (5)	& \bf{6.66} & \bf{ 0.20} & 1.37 & 
				88.0 &  94.4 &  95.9 : 92.1  &  
				88.7 &  93.4 &  92.4 : 94.8 \\ 
\RUE 			&  8.71 & 0.28 & 1.87 &
				85.0 & 91.3 & 93.8 : 87.8 &
				86.3 & 90.7 & 89.5 : 92.6				 
\\ \hline	
\MLE 			& 8.81 & 0.30 & 3.59 &  
				85.0   &  91.3  &  93.8 : 87.8 & 
				86.3 &  90.7 &  89.5 : 92.6 \\
\TCS   			&  8.81 & 0.26 & \bf{0.52} & 
				 89.0 &  95.3 &  96.5 : 93.4 &  
				 88.9 &  94.0 &  92.7 : 95.0 
\\  \hline
\MCD			& 8.83 &   0.26 &   0.58 &
				81.8 &  88.6 &  91.7 : 84.3 &
				86.0 &  91.6 &  90.1 : 94.0 \\
\BNNvi 	& 11.09 &   0.33 &   1.57 &
				79.9 &  87.3 &  90.5 : 83.3 & 
				85.7 &  91.4 &  89.5 : 93.9  \\
\emph{\textsf{BNN(LL-VI)}} 	&   8.94 &   0.33 &   4.15 &
				 84.6 &  91.0 &  93.5 : 87.2 & 
				87.8 &  93.3 &  91.7 : 95.4  \\
\KFAC   			&  8.75 & 0.29 & 3.45 &
				85.0 & 91.3 & 93.8 : 87.8 &
				86.3 & 90.7 & 89.5 : 92.6
\\ \hline
\rowstyle{\bfseries}
{\normalfont \mfzero} & {\normalfont 8.81} & {\normalfont 0.26} & {\normalfont 0.56} &  
  				91.0 &  96.4 &  97.4 {\normalfont:}  94.8 & 
				89.7 &  94.6 &  93.3 {\normalfont:}  95.3 \\ \hline
\end{tabular}
\end{adjustbox}
\end{table*}

\begin{table*}[th]
\centering
\caption{Out of Distribution Detection on CIFAR-100 versus SVHN.}\label{tab:tCifar100svhn}
\begin{tabular}{$cccc}
\hline
\multirow{2}{*}{Method} & \multicolumn{3}{c}{CIFAR-100 versus SVHN}\\\cline{2-4}
& Acc. & ROC & PR (in:out) \\
\hline
\ENS	 (5)	&      75.99 &  83.48 &  77.75 : 88.85 \\ 
\RUE$^\dagger$  &  73.90 &  80.69 &  74.03 : 86.93
\\ \hline
 \MLE 		&	73.90 &  80.69 &  74.03 : 86.93  \\
\TCS   	&	76.94 &  83.43 &  77.53 : 88.11 \\ \hline
\MCD		&	74.13 &  81.87 &  75.78 : 88.07 \\
\BNNvi$^\dagger$ &	71.87 &  79.55 &  72.08 : 87.27	 \\
 \KFAC  		& 74.13 &  80.97 &  74.46 : 87.06 \\ \hline
\mfzero &  \bf{81.38} &  \bf{88.04} &  \bf{84.59} {\normalfont:} \bf{91.23} \\ \hline
 \end{tabular}
 
 {$^\dagger$: Only the last layer is used  due to the high computational cost.}
\end{table*}

\begin{figure*}[t]
\centering
\subfigure[NLL $\color{blue}\downarrow$ on held-out]{%
\includegraphics[width=0.30\textwidth]{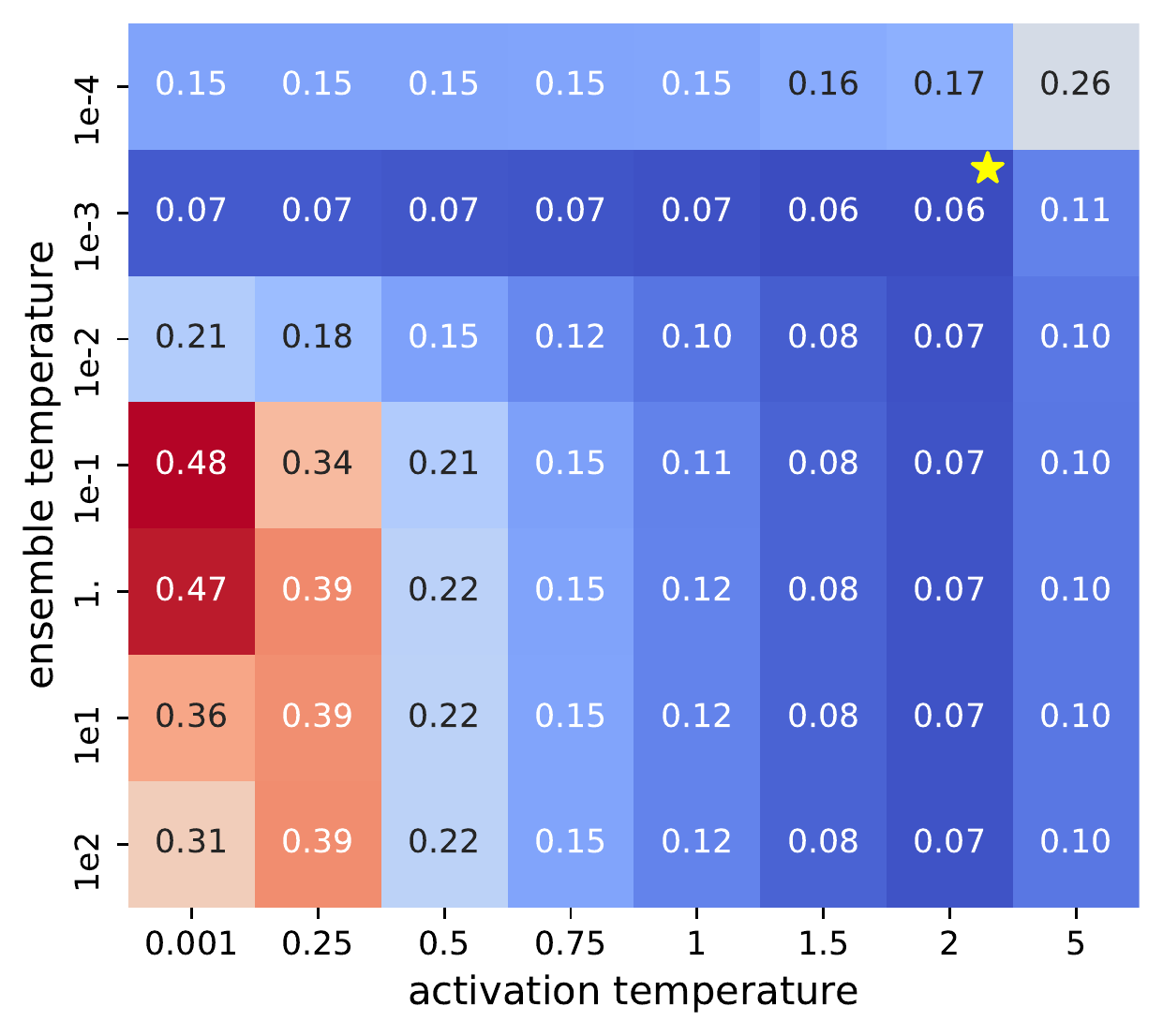}
\label{fig:subfigure1}}
\subfigure[auROC  $\color{red}\uparrow$ on in-/ood- held-out]{%
\includegraphics[width=0.30\textwidth]{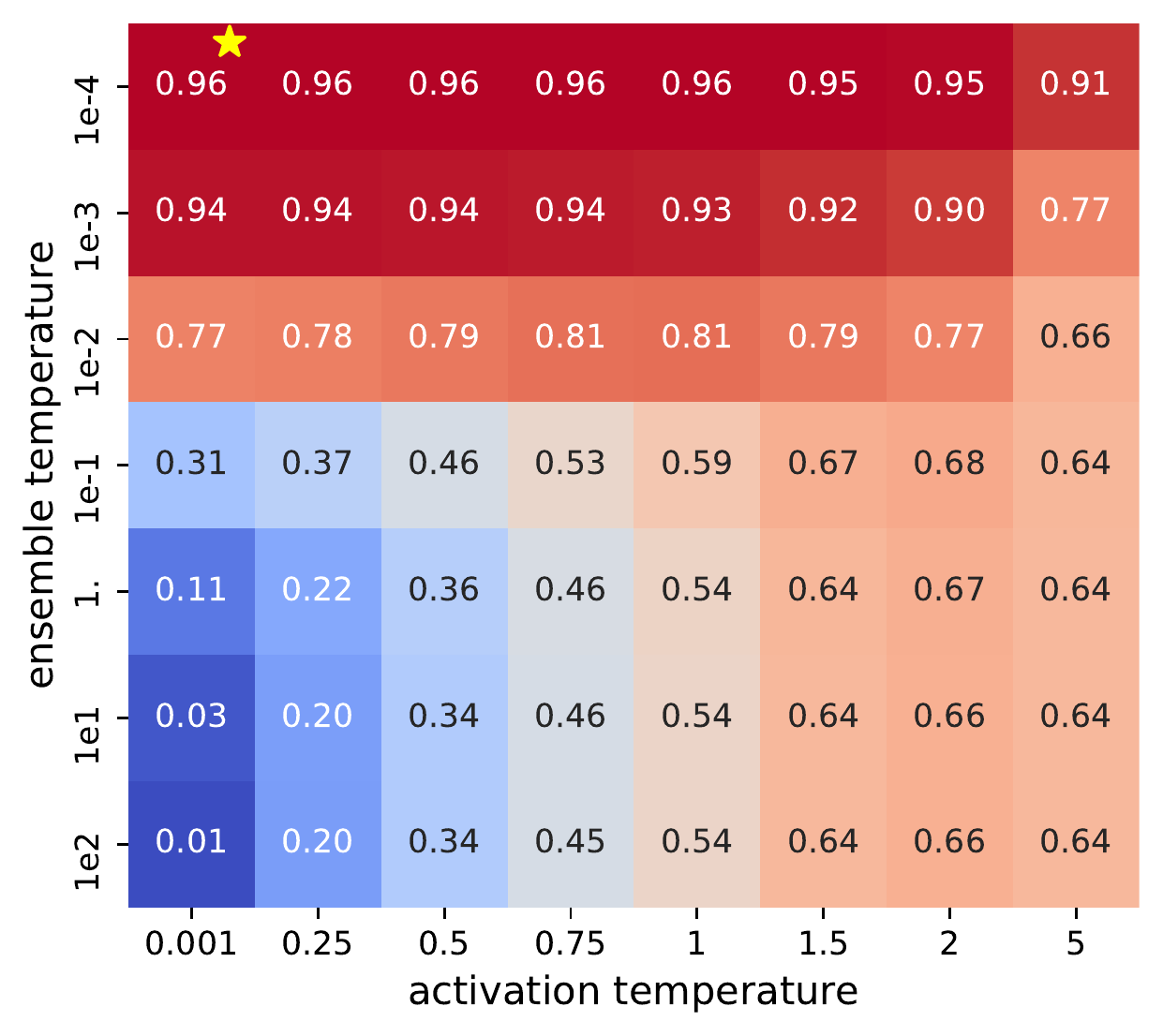}
\label{fig:subfigure2}}
\subfigure[Calibration under distribution shift]{%
\includegraphics[width=0.345\textwidth]{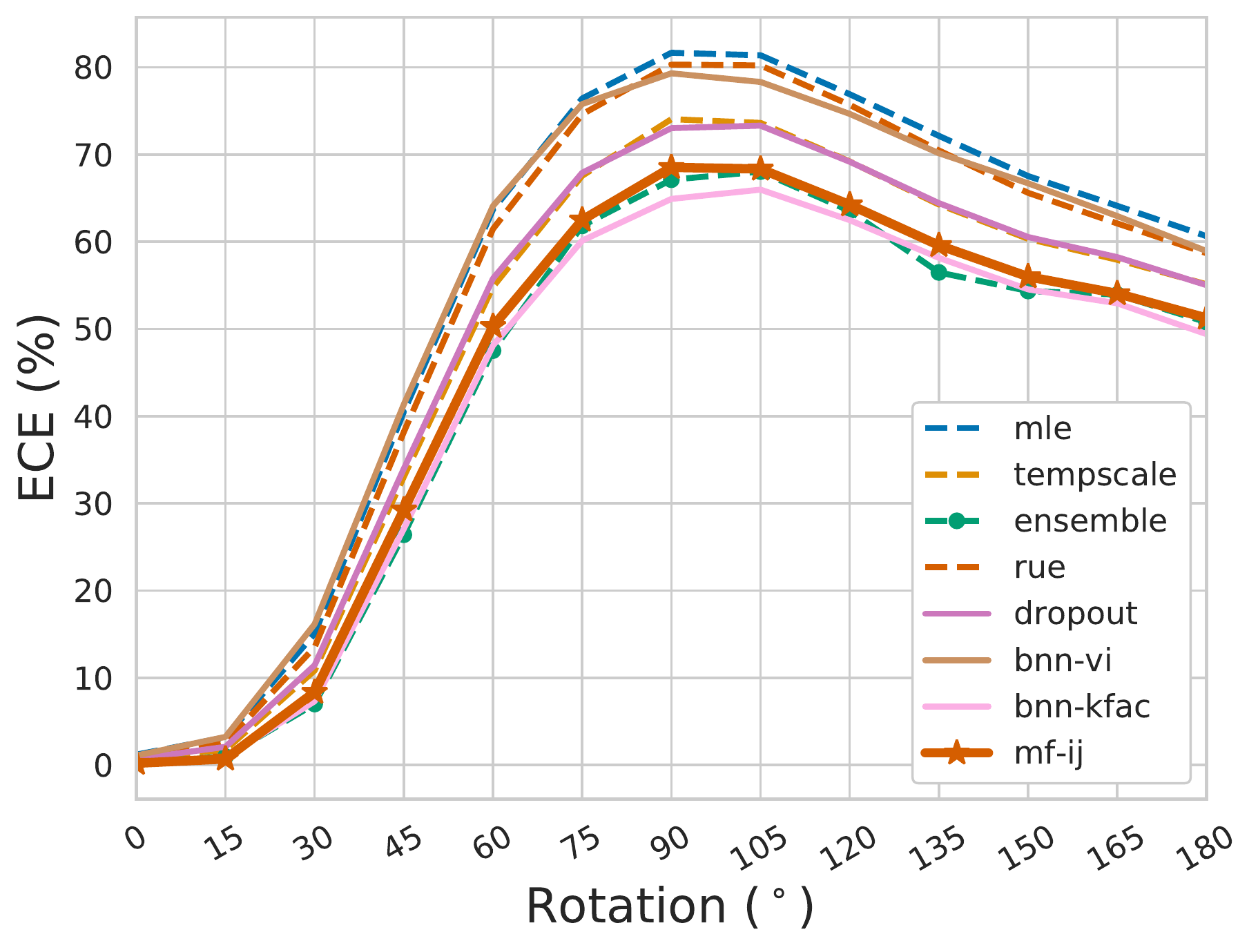}
\label{fig:subfigure3}}
\caption{Best viewed in colors. (a-b) Effects of the activation $\Tact$ and ensemble temperatures $\Tens$ on NLL  and AuROC. The yellow star marks the best pairs of temperatures. See \S\ref{ssec:temperature} for details. (c) \mfzero is better or as robust as other approaches under distributional shift. See \S\ref{ssec:shift} for details.}
\label{fig:fTempAct}
\end{figure*}

\subsection{The Effects of $\Tens$ and $\Tact$}\label{ssec:temperature}
We study the roles of ensemble and activation temperatures  in \mfgs. $\Tens$ controls the how wide we want our Gaussian is and thus determines how many models we would want to integrate. On the other hand, $\Tact$ controls how much we want each model in the ensemble to be confident about their predictions.

To this end, on MNIST and NotMNITST datasets we grid search the two temperatures and generate the heatmaps of NLL and AU ROC on the held-out sets, shown in ~\pref{fig:fTempAct}. Note that $(\Tens = \infty, \Tact=1)$ correspond to \MLE.    What is particularly interesting is that for NLL, higher activation temperature ($\Tact \ge1$) and lower ensemble temperature ($\Tens \le 10^{-2}$) work the best. For area under ROC, however,  lower temperatures on both work best. 
Using $\Tact>1$ for better calibration was also observed in~\citep{guo2017calibration}.  On the other end,  for OOD detection,  ~\citep{liang2017enhancing} suggests a very high activation temperature ($\Tact = 1000$ in their work, likely due to using a single model instead of an ensemble).

Examining both plots horizontally, we can see when the $\Tens$ is higher, the effect of $\Tact$ is more pronounced than when $\Tens$ is lower. Vertically, we can see when the $\Tact$ is lower, the effect of $\Tens$ is more pronounced than when $\Tact$ is higher, especially on the OOD tasks.

\subsection{Robustness to Distributional Shift.}\label{ssec:shift}
\cite{snoek2019can} points out that many uncertainty estimation methods are sensitive to distributional shift. Thus, we evaluate the robustness of \mfzero on rotated MNIST images, from $0$ to $180^{\circ}$. The  ECE curves in \pref{fig:fTempAct}(c) show \mfzero is better or as robust as other approaches.

\section{Unscented Kalman Filter for Gaussian-Softmax Integral} 
Unscented Kalman Filter (\ukf) is a standard numerical technique~\citep{wan2000unscented} used in nonlinear estimations to approximate Gaussian integrals. Here we briefly summarize the main steps in \ukf.  

\begin{table}[thbp]
\caption{\mfgs Compared to \ukf for Uncertainty Estimation.}
\label{tUKF}
\centering
\begin{tabular}{rccc||ccc}
\hline
 \multirow{3}{*}{ Method} &  \multicolumn{3}{c||}{MNIST } & \multicolumn{3}{c}{NotMNIST } \\ 
 & \multicolumn{3}{c||}{In-domain ($\downarrow)$ } & \multicolumn{3}{c}{OOD detection ($\uparrow$)}\\
\cline{2-7}
& $\varepsilon$ & NLL & ECE  & Acc. & ROC & PR(in:out) \\ \hline
\mfzero &  1.67 & 0.05 & 0.20 &  91.93 &  96.91 & 96.67 : 96.99  \\ 
\ukf &  1.67 & 0.05 & 0.42 & 89.07 & 96.37 & 95.32 : 97.42 \\
\hline
\end{tabular}
\end{table}

Assume the Gaussian random variable $\va$ is of $K$ dimensional.
\ukf approximates the Gaussian-Softmax integral (reiterate Eq.(6) in the main text) by a carefully designed weighted sum on $2K+1$ sample points, referred to as \emph{sigma points} $\{\cst{a}_i\}$. The sigma points $\{\cst{a}_i\}$ are chosen such that they capture the posterior mean and covariance accurately to the 3rd order Taylor series expansion for any nonlinearity.
\begin{align}
e_k   & = \int \textsc{softmax}_k(\va) \sN(\va; \vmu, \mS) \mathrm{d}\va \approx \sum_{i =0}^{2K} \cst{w}_i \ \textsc{softmax}_k(\cst{a}_i).
\label{eGaussianSoftmax}
\end{align}
The specific form of $\cst{w}_i$ and $\cst{a}_i$ are as follows,
\begin{align}
& \cst{a}_0 = \vmu,  \quad   \cst{w}_0 = - \frac{\alpha }{1 - \alpha}; &&  \\
& \cst{a}_i =  \vmu + \sqrt{(1-\alpha)K}  \mat{L}_i,   \cst{w}_i = \frac{1}{2(1-\alpha) K }, && \text{ for }  i=1, \ldots, K; \\
 & \cst{a}_i =  \vmu - \sqrt{(1-\alpha)K} \mat{L}_i,  \cst{w}_i = \frac{1}{2(1-\alpha) K }, && \text{ for }  i=K+1, \ldots, 2K;
 \end{align}
where $\mat{L}_i$ is the $i$-th column of $\mat{L}$, which is the square root of the covariance matrix $\mS$. We use the Cholesky factor as $\mat{L}$ and the constant $\alpha = 0.5$ in our implementation.

We plug in \ukf approximation and evaluate its performance on uncertainty estimation. We compare \ukf with \mfzero on MNIST and NotMNIST datasets in Table~\ref{tUKF}. We also tuned activation and ensemble temperatures for \ukf. \ukf performs competitively on OOD detection task, but not as well on in-domain ECE compared with \mfzero. 


\section{Experiment Details}
In this section, we provide experimental details for reproducibility.
\subsection{Datasets} The statistics of the benchmark datasets used in our experiment are summarized in Table~\ref{tDatasets}.
\begin{table}[tbhp]
\caption{Datasets Used in Experiments}
\label{tDatasets}
\centering
\small
\begin{tabular}{lll}
\hline
Dataset& \# of classes  & train/held-out/test \\
\hline
MNIST  & 10 & 55k/5k/10k \\ 
CIFAR-10  & 10 & 45k/5k/10k  \\
CIFAR-100  &100 & 45k/5k/10k\\ 
ILSVRC-2012  & 1,000 & 1,281k/25k/25k \\ \hline

 NotMNIST  & - & -/5k/13.7k \\
 LSUN (resized)  & - &  -/1k/9k\\
  SVHN  & - &  -/5k/21k\\
  Imagenet-O  & - & -/2k/-\\
  \hline
\end{tabular}
\end{table}

\subsection{Definitions of Evaluation Metrics}~\label{sMetrics}
\textbf{NLL} is defined as the $\textsf{KL}$-divergence between the data distribution and the model's predictive distribution,
\begin{equation}
\text{NLL} = - \log p_\theta(y|x)= - \sum_{c=1}^K y_c \log p_\theta(y=c | x),
\end{equation}
where $y_c$ is the one-hot embedding of the label.

\textbf{ECE} measures the discrepancy between the predicted probabilities and the empirical accuracy of a classifier in terms of $\ell_1$ distance. It is computed as the expected difference between per bucket confidence and per bucket accuracy. All predictions $\{p(x_i)\}_{i=1}^N$ are binned into $S$ buckets such that $B_s = \{ i \in [N] | p(x_i)  \in I_s\}$ are predictions falling within the interval $I_s$. ECE is defined as,
\begin{equation}
\text{ECE} = \sum_{s=1}^S \frac{|B_s|}{N}  \left| \text{conf}(B_s) -  \text{acc}(B_s)\right|, 
\end{equation}
where $\text{conf}(B_s) = \frac{\sum_{i \in B_s} p(x_i) }{|B_s|}$ and $\text{acc}(B_s) = \frac{\sum_{i \in B_s} \bm{1}\ [\hat{y}_i = y_i] }{|B_s|}$.

\subsection{Hyper-parameter Tuning}
\textbf{Hyper-parameters in training} \quad Table~\ref{tab:tHparams} provides key hyper-parameters used in training deep neural networks on different datasets.
\begin{table*}[th]
\centering
\caption{Hyper-parameters of Neural Network Trainings}\label{tab:tHparams}
\begin{tabular}{r|llll}
\hline
Dataset & MNIST & CIFAR-10 & CIFAR-100 & ImageNet \\ \hline
Architecture & MLP & ResNet20 & Densenet-BC-121 & ResNet50\\
Optimizer & Adam   & Adam       & SGD 			  & SGD \\
Learning rate & 0.001 & 0.001 	 & 0.1  			  & 0.1	\\
\multirow{2}{*}{Learning rate decay} 
			& exponential 	& 	staircase $\times 0.1$ & staircase $\times 0.1$ &  staircase $\times 0.1$\\
			&  $\times$ 0.998 & at 80, 120, 160 		  & at 150, 225 			& at 30, 60, 80\\
Weight decay & 0 & $1e^{-4}$   & $5e^{-4}$			& $1e^{-4}$ \\
Batch size & 100 & 8			& 128 			& 256 \\
Epochs & 100 & 200			& 30	 			& 90\\
\hline
\end{tabular}
\end{table*}

\textbf{Hyper-parameters in uncertainty estimation} \quad
For \mfzero method, we use $\lambda_0 = \frac{3}{\pi^2}$ in our implementation. 
For the \ENS approach, we use $M=5$ models on all datasets as in~\citep{snoek2019can}. For \RUE, \MCD, \BNNvi and \KFAC, where sampling is applied at inference time, we use $M=500$ Monte-Carlo samples on MNIST, and $M=50$ on CIFAR-10, CIFAR-100 and ImageNet. We use $B=10$ buckets when computing ECE on MNIST, CIFAR-10 and CIFAR-100, and $B=15$ on ImageNet.

For in-domain uncertainty estimation, we use the NLL on the held-out sets to tune hyper-parameters. For the OOD detection task, we use area under ROC curve on the held-out to select hyper-parameters. We report the results of the best hyper-parameters on the test sets. The key hyper-parameters we tune are the temperatures, regularization or prior in \BNN methods, and dropout rates in \MCD.

\textbf{Other implementation details}\quad   When Hessian needs to be inverted, we add  a
dampening term $\tilde{H} = (H + \epsilon I)$ following~\citep{ritter2018scalable,schulam2019can} to ensure positive semi-definiteness and the smallest eigenvalue of  $\tilde{H}$ be 1. For  \BNNvi, we use Flipout~\citep{wen2018flipout} to reduce gradient variances and  follow~\citep{snoek2019can} for variational inference on deep ResNets.
On ImageNet, we compute the Kronecker-product factorized Hessian matrix, rather than full due to high dimensionality. For \KFAC and \RUE, we use mini-batch approximations on subsets of the training set to scale up on ImageNet, as suggested in~\citep{ritter2018scalable,schulam2019can}.

\section{Infinitesimal Jackknife and Its Distribution}
\label{sJackknife}

We include a derivation of \ijknife (Eq.(26) in \S4.3 of the main text) with more details to be self-contained.

Jackknife is a resampling method to estimate the confidence interval of an estimator~\citep{tukey1958bias,efron1981jackknife}. It works as follows: each element $z_i$ is left out from the dataset $\sD$ to form a unique ``leave-one-out'' jackknife sample $\sD_i = \sD - \{z_i\}$. A jackknife sample's estimate of  $\theta$ is given by
\begin{equation}
\ttheta_i = \argmin_\theta \sL(\sD_i; \theta)
\end{equation}
We obtain $N$ such samples $\{\ttheta_i\}_{i=1}^N$ and use them  to estimate the variances of $\htheta_N$ and the predictions made with $\htheta_N$. In this vein, this is a form of ensemble method.

However, it is not feasible to retrain modern neural networks $N$ times, when $N$ is often in the order of millions. \IJknife is a classical tool to approximate $\ttheta_i$ without re-training on $\sD_i$. It is often used as a theoretical tool for asymptotic analysis~\citep{jaeckel1972infinitesimal}, and is closely related to influence functions in robust statistics~\citep{cook1982residuals}.  Recent studies have brought (renewed) interests in applying this methodology to machine learning problems~\citep{giordano2018swiss, koh2017understanding}. Here, we briefly summarize the method. 

\textbf{Linear approximation.}\quad The basic idea behind \ijknife is to treat the $\htheta_N$ and $\ttheta_i$ as special cases of an estimator on weighted samples,
\begin{equation}
\htheta(\vw) =  \argmin_\theta \frac{1}{N}\sum\nolimits_i w_i \ell( z_i; \theta)\footnote{The loss in Eq.(\ref{eIJloss}) is scaled by $\frac{1}{N}$ compared to Eq.(2) in the main text. Nonetheless, the minimizer $\htheta(\vw)$ is the same.}, \quad w_i \geq 0.\label{eIJloss}
\end{equation}
Thus the maximum likelihood estimate $\htheta_N$ is $\htheta(\vw)$ when $w_i = 1, \forall i$. A jackknife sample's estimate $\ttheta_i$, on the other end, is $\htheta(\vw)$ when $\vw = \ones - \ve_i$ where $\ve_i$ is an all-zero vector except taking a value of $1$ at the $i$-th coordinate. 

Using the first-order Taylor expansion of $\htheta(\vw)$ around $\vw =\ones$, we obtain (under the condition of twice-differentiability and the invertibility of the Hessian),
\begin{equation}
\ttheta_i \approx \htheta_N + H_N^{-1} \nabla \ell(z_i, \htheta_N) 
\stackrel{\text{def}}{=} \htheta_N +  H_N^{-1}\nabla \ell_i \label{ePerturb}
\end{equation}
where $H_N = \sum_i \nabla \nabla\T \ell_i$ is the Hessian matrix of $\sL$ evaluated at $\htheta_N$(recall Eq.(18) in the main text), and $ \nabla \ell_i$ is the gradient of $\ell(z_i; \theta)$ evaluated at $\htheta_N$. 

\textbf{An infinite number of \ijknife samples.} If the number of samples $N\rightarrow +\infty$, we can characterize the ``infinite'' number of $\ttheta_i$ with a closed-form Gaussian distribution $\ttheta_i \sim \sN(\vmu_N, \mSigma_N)$ with the following sample mean and covariance as the distribution's mean and covariance,
\begin{align}
\vmu_N & = \frac{1}{N} \sum_i \ttheta_i = \htheta_N + \frac{1}{N} H_N^{-1}  \sum_i \nabla \ell_i(\htheta_N)  =  \htheta_N, \label{eDistribution}
\end{align}
and
\begin{align}
\mSigma_N & = \frac{1}{N} \sum_i(\ttheta_i-\htheta_N)(\ttheta_i-\htheta_N)\T = H_N^{-1}\left[ \frac{1}{N}\sum_i \nabla \ell_i\nabla \ell_i\T\right]H_N^{-1} = \frac{1}{N}  H_N^{-1}J_NH_N^{-1} 
\end{align}
where $J_N = \sum_i \nabla \ell_i\nabla \ell_i\T$ is the \emph{observed} Fisher information matrix (recall Eq.(18) in the main text). 

\textbf{An infinite number of infinitesimal bootstrap samples.}\quad  The above procedure and analysis can be extended to bootstrapping (\ie, sampling with replacement). The perturbed estimates $\ttheta_b$ from bootstrap samples have
\begin{equation}
\ttheta_b \approx  \htheta_N +  \sum_i  \omega^b_i H_N^{-1}  \nabla \ell_i, \text{ where } \vomg_b \sim \text{Multinomial} (N, \frac{1}{N}) \label{ePerturb2}.
\end{equation} 
Similarly, to characterize the estimates from the bootstraps, we can also use a Gaussian distribution,
\begin{equation}
\ttheta_b \sim \sN\left(\htheta_N, \mSigma_N \right),  \quad \text{with }  \mSigma_N =  H_N^{-1}J_NH_N^{-1}.\label{eDistribution2}
\end{equation}
\citet{lakshminarayanan2017simple} discussed that using models trained on bootstrapped samples does not work empirically as well as other approaches, as the learner only sees $\sim63\%$ of the dataset in each bootstrap sample. 
We note that this is an empirical limitation rather than a theoretical one. In practice, we can only train a very limited number of models. However, we hypothesize that we can get the benefits of combining an \emph{infinite} number of models from the closed-form Gaussian distribution \emph{without} training. Empirical results valid this hypothesis.


\vskip 0.2in
\bibliography{uncertainty}

\end{document}